\documentclass{article} 
\usepackage{iclr2021_conference}
\usepackage{times}
\usepackage{enumerate}
\usepackage{caption}

\usepackage{amsmath,amsfonts,bm}

\def\eqref#1{equation~\ref{#1}}

\def\1{\bm{1}}

\DeclareMathAlphabet{\mathsfit}{\encodingdefault}{\sfdefault}{m}{sl}
\SetMathAlphabet{\mathsfit}{bold}{\encodingdefault}{\sfdefault}{bx}{n}

\usepackage{booktabs}
\usepackage{hyperref}
\usepackage{multirow}
\usepackage{graphicx}
\usepackage{url}

\title{Unsupervised Word Translation Pairing using Refinement based Point Set Registration}

\author{Silviu Oprea \thanks{The work was performed while the author was at Huawei Ireland Research Centre, Dublin.} \\
University of Edinburgh \\
Edinburgh, UK \\
\texttt{silviu.oprea@ed.ac.uk} \\
\And
Sourav Dutta \\
Huawei Research Centre \\
Dublin, Ireland \\
\texttt{sourav.dutta2@huawei.com} \\
\And
Haytham Assem \\
Huawei Research Centre \\
Dublin, Ireland \\
\texttt{haytham.assem@huawei.com} \\
}

\newcommand{\todo}[1]{\bgroup\color{red}#1\egroup\marginpar{\color{red}TODO}}

\newcommand{\actg}{\emph{BioSpere}}
\renewcommand{\cite}{\citep}
\newcommand{\comment}[1]{}

\iclrfinalcopy

\begin{document}

\maketitle

\begin{abstract}
\vspace*{-1mm}
    Cross-lingual alignment of word embeddings play an important role in knowledge transfer across languages, for improving
    machine translation and other multi-lingual applications. Current unsupervised approaches rely on similarities in geometric
    structure of word embedding spaces across languages, to learn structure-preserving linear transformations using adversarial
    networks and refinement strategies. However, such techniques, in practice, tend to suffer from instability and convergence
    issues, requiring tedious fine-tuning for precise parameter setting. This paper proposes $\actg$, a novel framework for
    unsupervised mapping of bi-lingual word embeddings onto a shared vector space, by combining {\em adversarial initialization}
    and {\em refinement procedure} with {\em point set registration} algorithm used in image processing. We show that our framework
    alleviates the shortcomings of existing methodologies, and is relatively invariant to variable adversarial learning performance,
    depicting robustness in terms of parameter choices and training losses. Experimental evaluation on parallel dictionary induction task demonstrates state-of-the-art results for our framework on diverse language pairs.
    \comment{
    Cross-lingual word alignment plays an important role in the construction of parallel dictionaries for machine translation
    and transfer learning applications. Current unsupervised approaches rely on learning transformations using refinement
    procedures and/or adversarial training for mapping monolingual word embeddings onto a shared vector space. However, such
    techniques, in realistic scenarios, tend to suffer from stability and convergence issues unless favourable evaluation
    conditions, in terms of precise parameter setting and language selection, are provided.
    This paper proposes a novel framework for unsupervised mapping of bi-lingual vocabulary into a common space, by combining
    {\em adversarial training} and {\em refinement procedure} with {\em point set registration} approach. We demonstrate that
    this coupling with image processing techniques, like CycleGAN and Coherent Point Drift, not only captures best-of-the-worlds
    but also alleviates the shortcomings of existing methods, making it robust in terms of parameter choices and training losses
    and convergence. Extensive experiments on diverse languages for parallel dictionary construction
    task showcase the efficacy and robustness of our framework over state-of-the-art strategies.
    }
\vspace*{-2mm}
\end{abstract}

\section{Introduction and Background}
\label{sec:intro}
\vspace*{-2mm}

With the success of {\em distributed word representation}, like Word2Vec~\cite{w2v}, GloVe~\cite{glove} and FastText~\cite{fasttext}, in capturing 
rich semantic meaning, the use of these embeddings has permeated a wide range of Natural Language Processing (NLP) tasks such as text classification, 
document clustering, text summarization and question answering~\cite{klem12} to name a few. Unsupervised learning of such continuous high dimensional 
vector representation for words rely on the {\em distributional hypothesis}~\cite{harris}. 

{\bf Motivation.} As a natural generalization, methods for obtaining multi-lingual word embeddings across diverse languages have recently gained significant attention in 
the NLP research community~\cite{joint}. Learning {\em cross-lingual word embeddings} (CLWE) entails mapping the vocabularies of different languages onto a single vector 
space for capturing syntactic and semantic similarity of words across languages boundaries~\cite{upa16}. Thus, CLWE provides an effective approach for knowledge 
transfer across languages for several downstream linguistics tasks such as machine translation~\cite{vecmap, lample18a, lample18b}, POS tagging~\cite{zh16}, dependency parsing~\cite{dp},
named entity recognition~\cite{xie18, chen19}, entity linking~\cite{roth16}, language inference~\cite{conneau18} and low-resource language understanding~\cite{guo14}. 
In fact, word alignment across languages also finds interesting applications in the study of cultural connotations~\cite{culture} and spatio-linguistic commonalities~\cite{prep, cogn, typo}.

{\bf Linguistic Correlation.} Monolingual representation spaces learnt independently for different languages tend to exhibit similarity in terms of {\em geometric properties 
and orientations}~\cite{langsim}. For example, the vector distribution of numbers and animals in English show a similar geometric constellation formation as their Spanish counterparts.
Further, the frequency of words across languages have been shown to follow the {\em Zipf's distribution}~\footnote{\scriptsize observed on 10 million words from Wikipages across 30 languages 
as shown in \url{en.wikipedia.org/wiki/Zipf's_law}}, with nearly $70\%$ most frequent word overlap~\cite{ssim} and $60\%$ synonym overlap~\cite{syno} across language pairs. 
Existing techniques for extracting cross-lingual word correspondences rely on above inter-dependencies to efficiently learn transformations across the monolingual embedding spaces.

{\bf State-of-the-art \& Challenges.} Early approaches for directly obtaining multi-lingual word embeddings relied on the availability of large parallel corpora~\cite{guow15} or document-aligned comparable 
corpora~\cite{mog16, vul16}. However, such methods are not scalable as annotations are expensive and large parallel datasets, especially for low-resource languages, are scarce in practice. To address the 
above challenges, linear transformations between two monolingual embedding space using small manually created bi-lingual dictionaries were proposed~\cite{langsim, vecmap-sup}. Words having similar surface 
forms across languages were used to induce seed dictionaries and other augmented refinement strategies were explored in the semi-supervised approaches of \citet{vecmap-semi, dens, middle}. Subsequently, 
improvements in orthogonality and optimization constraints were explored for generalization beyond bi-lingual settings for supervised cross-lingual alignment and joint training methods~\cite{rcsls, geomm, umh, joint}.

Unsupervised framework for bi-lingual word alignment was first proposed by \citet{val16, zh17a, zh17b} using {\em adversarial training}. The use of post-mapping refinements were shown to produce high quality 
results in the MUSE framework~\cite{muse} across diverse languages, and was used for machine translation system in~\cite{lample18a, lample18b}. Parallel dictionary construction using {\em CSLS}~\cite{muse} 
(adopted in this paper) or inverted softmax~\cite{isf} was shown to tackle the ``hubness problem''~\cite{hub} caused due to highly dense vector space regions (called {\em hubs}), which 
adversely affects reliable retrieval of bi-lingual word translation pairs. However, the performance of adversarial learning techniques have been shown to suffer from instability, convergence issues, 
and dependence of precise parameter settings. Further, \citet{limi} found the above unsupervised approaches to fail for morphologically rich languages. Hence, optimization formulations 
using Gromov-Wasserstein, Sinkhorn distance, and Iterative Closest Point were explored~\cite{wsproc, gwproc, sinkhorn, nonadv}. Recently, {\em adversarial auto-encoders} using 
{\em cyclic loss optimization} in latent space supplemented with refinements~\cite{advauto, coli2020} has achieved state-of-the-art results for bi-lingual word embedding alignment on diverse languages.

\vspace*{1mm}
{\bf Proposed Approach.} In this paper, we propose $\actg$ (\underline{B}i-L\underline{i}ngual W\underline{o}rd Tran\underline{s}lation via \underline{P}oint S\underline{e}t \underline{Re}gistration), a novel framework for 
{\em fully unsupervised bi-lingual word correspondence induction}. Given two independently learnt monolingual word embedding space, $\actg$ uses a combination of adversarial training, refinement procedure, 
and point set registration approach to efficiently extract word translations. Specifically, the input vector spaces are initially aligned using {\em CycleGAN}, a Generative Adversarial Network (GAN) 
trained using {\em cycle-consistency loss} optimization criteria, as word translation pairs are {\em symmetric}, i.e., if word $w_x$ is a translation of $w_y$, then $w_y$ is also a translation of $w_x$. The cyclic loss 
criteria has been shown to be better in capturing bi-directional distributional similarities~\cite{sinkhorn} and in training adversarial networks in \cite{coli2020} (auto-encoders with a latent space of the embeddings). 
The word alignments obtained 
from CycleGAN are then refined via {\em symmetric re-whitening} or spherical transformation~\cite{white} to remove correlations among the different components of the language embeddings. It is interesting to 
note that extracting {\em word correspondences is akin to point set registration}~\cite{psr} in image processing. To this end, $\actg$ finally utilized the {\em Coherent Point Drift} (CPD) algorithm~\cite{cpd} 
to compute an affine transformation between the aligned and refined vector spaces. Our choice of CPD hinges on two key insights: {\bf (i)} CPD inherently works on the concept of {\em Gaussian Mixture Model} (GMM), 
which has been shown to tackle the {\em ``hubness'' problem}~\cite{dens}; and {\bf (ii)} CPD being an unsupervised approach might reduce error propagation from the adversarial or refinement steps, as opposed to the 
supervised Procrustes refinement~\cite{proc} (extensively used in the literature) that requires an intermediate synthetic (possibly erroneous) dictionary creation from the adversarial training stage. Extensive 
empirical results on diverse languages (reported in Section~\ref{sec:expt}) demonstrate that the proposed $\actg$ framework outperforms existing approaches in terms of accuracy for parallel dictionary creation. We further show that $\actg$ can robustly handle adversarial convergences issues, sub-optimal parameter settings, as well as morphologically rich and low-resource languages.

\vspace*{2mm}
{\bf Contributions.} In a nutshell, the key contributions of this paper can be described as follows:
\begin{itemize}
	\item $\actg$, an {\em unsupervised} framework for learning bi-lingual word translations from two independent monolingual embedding spaces -- thus aligning the vocabularies to a common vector representation 
	for capturing semantic similarities between words across languages;
	\item A novel combination of adversarial training, refinement procedure, and point set registration algorithm -- coupling the advantages of {\em cycle-consistence loss} and {\em Gaussian Mixture Model} -- to alleviate 
	the challenges for word embedding space alignment;
	\item Unsupervised stopping criterion incorporating {\em cycle-loss consistency} measure, with better correlation with mapping quality, for selection of adversarial training model parameters;
	\item Experimental evaluation on diverse language pairs showcasing {\em enhanced accuracy} (nearly at par with supervised approaches) compared to existing techniques, for parallel dictionary construction  task, even for small vocabulary sizes; and,
	\item {\em Robustness} study of $\actg$ framework in efficiently handling hubness problem, dependencies on adversarial learning convergence and precise parameter choice, as well as morphologically rich or low-resourced languages.
\end{itemize}


\section{Framework}
\label{sec:frame}

\begin{figure*}[t]
\centering
\vspace*{-7mm}
	\includegraphics[width=\columnwidth]{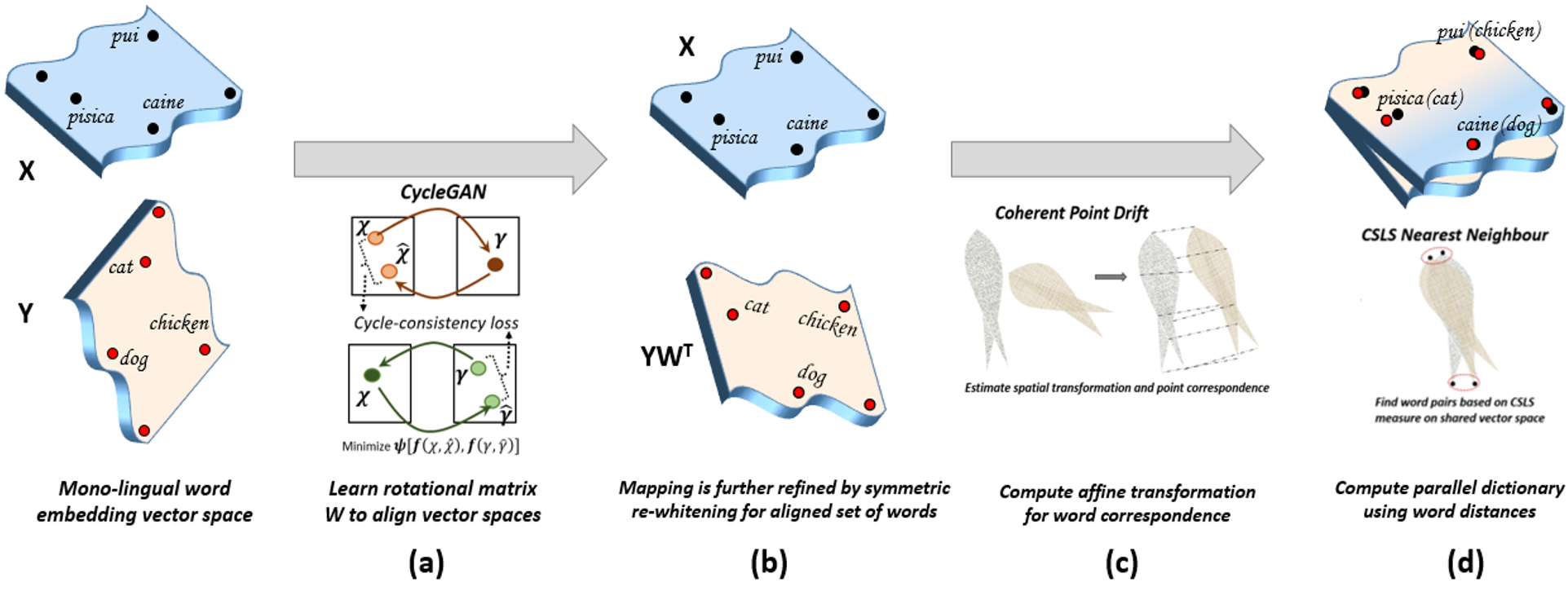}
	\vspace*{-7mm}
	\caption{Toy illustration (on {\em en-ro} language pair) of the different modules of $\actg$ -- (a) {\em Align}, (b) {\em Correspond}, (c) {\em Transform}, and (d) {\em Generate} --  for unsupervised 
	parallel dictionary construction.}
	\label{fig:actg}
\end{figure*}
We assume the existence of two sets $X=\{x_n\}_{n=1}^N$ and $Y=\{y_m\}_{m=1}^M$ of word embeddings trained independently on monolingual data from a source and a target language, respectively. The aim of our $\actg$ framework is to map each word in the source language to its translation in the target language, in a manner that does not require any cross-lingual supervision. Equivalently, we wish to align the two embedding sets in such a way that words that are semantically similar across languages are close to each other.

To achieve this, we hinge on $4$ modules, namely {\em Align, Correspond, Transform and Generate} ({\bf ACTG})~\footnote{Inspired by the $4$ bases (Adenine, Cytosine, Thymine \& Guanine) 
in DNA, the building block of life.} A pictorial depiction of the overview of the functionality of the different modules is presented in Figure~\ref{fig:actg}. We now look at each module individually.

\subsection{Align}
Our first module estimates an initial mapping using a domain-adversarial approach~\citep{dagan:gan:2016}. Let $x\sim p_{data}(x)$ and $y\sim p_{data}(y)$ be the data distributions. We learn two linear mappings $F:X\rightarrow Y$ and $G:Y\rightarrow X$, that we refer to as forward and backward \emph{generators}, respectively. We then train a model $D_Y$ to discriminate between synthetic target embeddings $Y_{syn}=F X=\{F x_n\}_{n=1}^N$, and real ones $Y$. Similarly, we train $D_X$ to discriminate between synthetic source embeddings $X_{syn}=G Y=\{G y_m\}_{m=1^M}$ and $X$. Note the notation overloading: we have used $F$ and $G$ to refer both to the parametric linear operators, as well as to the matrices of their parameters. We continue this way for simplicity, unless the context makes the reference ambiguous.

This results in a two-player game, where the discriminators aim to distinguish real and synthetic embeddings, while the generators aim at making their image as close to their codomain as possible, prevent discriminators from making accurate predictions.

We resemble this game in our training objective, which includes two categories of terms. The \emph{adversarial loss}, formulated for matching the distribution of the synthetic embeddings to the real distribution. For the forward generator $F:X\rightarrow Y$, and its corresponding discriminator $D_Y$, our adversarial loss is:
\begin{equation}
    \mathcal{L}_{adv}(F, D_Y, X, Y)=\mathbb{E}_{y\sim p_{data}(y)}[\log D_Y(y)] + \mathbb{E}_{x\sim p_{data}(x)}[\log (1 - D_Y(F(x))]
\end{equation}
We use a similar adversarial loss $\mathcal{L}_{adv}(G, D_X, Y, X)$ for the backward generator $G:Y\rightarrow X$ and its corresponding discriminator $D_X$.

The second objective category is in line with the work of \citet{coli2020}. Similar to them, we note that an adversarial generator could map the same set of source embeddings to any random permutation of target embeddings, as long as the synthetic distribution matches the target distribution. To account for this possibility, we argue that the learned generators should not contradict each other, but should be cycle-consistent. That is, given a source embedding $x$, the forward translation cycle should attempt to produce an output that coincides with $x$, i.e. $G(F(x))\approx x$. Analogously for the backward translation cycle, $G(F(y))\approx y$. We capture this endeavour with the addition of a \emph{cyclic loss} to our objective:
\begin{equation}
    L_{cyc}(F, G)=\mathbb{E}_{x\sim_{data}(x)}\left\lVert G(F(x)) \right\rVert_2 + \left\lVert F(G(y)) \right\rVert_2
\end{equation}
Following \citet{muse}, we make sure $F$ and $G$ remain roughly orthogonal during training by alternating model parameter update with $F\leftarrow(1+\beta)F - \beta(FF^T)F$, proceeding analogously for $G$. Intuitively, this preserves the monolingual quality of our embeddings by preserving their dot product and $l_2$ distances.

The output of this module are the two sets $X_A=F(X)$ and $Y_A=G(Y)$ of aligned embeddings (the images of the learned transformations).
\subsection{Correspond}
\label{section:frame:subsection:correspond}
Our vanilla CPD results, despite better than previous adversarial networks, are not au par with supervised work. To address this, we perform a set of refinement steps. In Correspond, the first refinement module, we perform symmetric re-weighting, successfully applied in previous work for word embedding alignment refinement\citep{vecmap,vecmap-sup,vecmap-semi,coli2020}. This requires a seed parallel dictionary. We induce such a dictionary by considering mutual nearest neighbours across the the original and mapped embeddings in both directions. That is, given mappings $f:X\rightarrow Y$ and $g:Y\rightarrow X$, the similarity between $x_n$ and $y_m$ is $\sigma_{nm}=\delta(f(x_n), y_m)+\delta(x_n, g(y_m))$, where $\delta$ is a metric in both $X$ and $Y$. Our metric of choice is cross-domain similarity local scaling (CSLS) \citep{muse}, shown by \citep{muse} to effectively address the hubbness problem, stereotypical especially when working in high-dimensional spaces. Using the bidirectional nature of our adversarial network when computing the similarity has not been done in previous, work to our knowledge, and we found it to considerably improve word translation performance. During dictionary induction, we only consider the 25K most frequent words from the source and target languages.
 
In the first step of this module we length-normalise and mean-center $X$ and $Y$, then apply a linear transformation with corresponding whitening matrices $W_x=(X^T X)^{-1/2}$ and $W_y=(Y^T Y)^{-1/2}$, i.e. $X_w = X W_x$ and $Y_w = Y W_y$. This makes the embedding dimensions ucorrelated among themselves.

Next, let $X^d$ and $Y^d$ be two matrices that reflect our seed dictionary, with $X^d_i$ being the embedding of a source word $Y^d_i$ being the embedding of its translation.
We perform an orthogonal transformation with symmetric re-weighting. Specifically, we compute $X_o=X_w U S^{1/2}$ and $Y_o=Y_w V S^{1/2}$ where $U$, $S$, and $V$ come from the singular value decomposition $USV^T=(X^d_w)^T Y^d_w$. This transposes the source and target embeddings into a common vector space.

In a final step, we perform de-whitening, to restore the original covariance in the embedding dimension distributions. That is, this module outputs $X_C=X_o U^T(X^T X)^{1/2}U$ and $Y_C=Y_o V^T(Y^TY)^{1/2}V$.
\subsection{Transform}
In this module we perform a further refinement of the transformed embeddings $X_C$ and $Y_C$ using affine Coherent Point Drift (CPD), a probabilistic framework suggested by~\citet{cpd} to perform point set registration, particularly in computer vision applications. The main idea is to consider the task of aligning the two embedding sets as a density estimation problem, where one set represents Gaussian mixture model (GMM) centroids, and the other the data points. With the two sets aligned, word translations can be obtained using the maximum of the GMM posterior probability, given a source embedding.
Specifically, we consider the embeddings in $Y_C$ as GMM centroids and the ones in $X_C$ as data points, generated by the GMM. The GMM density has has the form:
\begin{equation}
    p(x)=\sum_{m=1}^{M+1}p(m)p(x|m)
\end{equation}
where
$p(x|m)=\frac{1}{(2\pi\sigma^2)^{D/2}} \exp{\left(\frac{\lVert x-y_m\rVert_2}{2\sigma^2}\right)}$
and $x \in X_C$, $y_m\in Y_C$.
We also add a uniform distribution $p(x|M+1)=1/N$ to account for outliers, resulting in a Uniform-Gaussian mixture model. Following the authors, we use equal isotropic covariances $\sigma^2$, and equal membership probabilities $P(m)=1/M$ for all GMM components.
We estimate the GMM centroid locations $\theta$ by minimising the negative log-likelihood function:
\begin{equation}
    L(\theta, \sigma^2) = - \sum_{n=1}^N \log \sum_{m=1}^M P(m)p(x|m).
\end{equation}
We use the Expectation Maximization (EM) algorithm~\citep{em-algo}. to find the parameters $\theta$ and $\sigma^2$. We direct the interested reader to a more detailed description of CPD provided by its original authors\citep{cpd}.

We use the affine version of CPD, which provides a tuple $(R, t, s)$, where $R$ is a rotation matrix, $t$ is a translation vector, and $s$ is a scaling constant. The transformed source embedding set is computed as $X_T=(R X_C^T * s + t)^T$. We run CPD twice for each language pair, once in each directions, providing us with $X_T$ and $Y_T$.
\subsection{Generate}
We iterate between Correspond and Transform modules until an model selection criterion degrades for two consecutive iterations. The criterion is specified in Section~\ref{section:frame:subsection:criteria}. Equipped with the final $X_T$ and $Y_T$, we compute the final estimated parallel dictionary using the same procedure as in Section~\ref{section:frame:subsection:correspond}. We compare this with ground truth parallel dictionaries to compute word translation accuracy.
\subsection{Unsupervised Model Selection}
\label{section:frame:subsection:criteria}
Being in an unsupervised setting, we cannot use a validation set to direct us in choosing the best performing setting of our framework.
We follow approaches suggested in previous work to address this issue, that we adapt to our framework.
We follow \citet{muse} in considering the closeness of the source and target mapped embedding spaces. Specifically, we consider the 25K most frequent source words, use CSLS to generate a translation for each of them, and compute the average cosine similarity between these pairs. In our scenario, we consider similarity in both the source and target spaces, as specified in Section~\ref{section:frame:subsection:correspond}, criterion that we found to be better linked to word translation accuracy, compared to the unidirectional setting used in previous work~\citep{muse,coli2020}.
\section{Empirical Evaluation}
\label{sec:expt}
\vspace*{-2mm}

In this section, we evaluate the performance of the proposed $\actg$ framework in mapping the input word embeddings onto a shared vector space, such that semantically similar words across languages 
are close to each other (in terms of distance) in the common space. We benchmark the accuracy of $\actg$ against several existing approaches on the tasks of {\em bi-lingual dictionary induction} and 
{\em sentence translation retrieval} across a diverse set of languages.

\subsection{Experimental Setup}
\label{ssec:setup}

{\bf Dataset.} Our experimental setup closely follows that of \citet{muse}, extensively used in the literature. As input vocabulary, we use the FastText monolingual vector embeddings (with a dimensionality 
of $300$)~\cite{fasttext} of the top $200K$ most frequent words in each language. We consider {\em seven} different language pairs including morphologically rich and low-resourced languages. Specifically, 
we use English (en), German (de), French (fr), Spanish (es), Russian (ru), Hebrew (he), Finnish (fi), and Romanian (ro) -- a diverse mix of {\em isolating, fusional and agglutinative 
language} with {\em dependent and mixed marking} as reported in~\citet{limi}. 

{\bf Evaluation.} We report the {\em Precision@1} (P@1) accuracy scores based on the CSLS criteria~\cite{muse} for our empirical evaluations. In the {\em word translation task}, we use the gold dictionary with 
1,500 source test words (and full 200K target vocabulary) for different language pairs (obtained from {\small \url{github.com/facebookresearch/MUSE}}). 
We also perform the above evaluations with a smaller input vocabulary, 
to simulate scenarios of limited domain-specific resources.

\comment{
While for {\em sentence translation 
retrieval}, we consider the Europarl corpus with 2,000 source sentence queries and 200K target sentences for each of the language pairs. 
}

{\bf Baselines.} The performance of $\actg$ is compared with the following {\em unsupervised} approaches: \\
{\em $~~$ (1) MUSE}~\cite{muse} -- GAN~\cite{gan} trained for extracting a synthetic parallel dictionary to learn transformations via Procrustes refinement~\cite{proc}~\footnote{\scriptsize Code available at 
\url{github.com/facebookresearch/MUSE}}; \\ 
{\em $~~$ (2) Adv-Auto}~\cite{coli2020} --  Current state-of-the-art using adversarial auto-encoder to create synthetic dictionary, which is refined by symmetric re-whitening \& Procrustes strategies~\footnote{\scriptsize Code 
from \url{ntunlpsg.github.io/project/unsup-word-translation} is updated as in~\citet{coli2020}}; \\
{\em $~~$ (3) VecMap}~\cite{vecmap} -- Robust self-learning iterative algorithms exploiting structural similarities between embedding spaces for alignment~\footnote{\scriptsize Code obtained from \url{github.com/artetxem/vecmap}}; \\ 
{\em $~~$ (4) SinkHorn}~\cite{sinkhorn}: GAN trained using a combination of cyclic consistency loss and Sinkhorn distance~\cite{sink} as objective function; \\ 
{\em $~~$ (5) Non-Adv}~\cite{nonadv} -- Proposes an alternative approach using dimensionality reduction with Iterative Closest Point~\cite{icp} algorithm to find word correspondences; \\
{\em $~~$ (6) Was-Proc}~\cite{wsproc} -- A bi-stochastic matrix is computed using the assignment problem by jointly optimizing Wasserstein distance~\cite{otp} and Procrustes transformation; \\
{\em $~~$ (7) GW-Proc}~\cite{gwproc} -- Word translation is formulated as an optimal flow problem across different domains using Gromov-Wasserstein distance~\cite{otp}; and \\
{\em $~~$ (8) UMH}~\cite{umh} -- Uses correlation between multiple languages for jointly learning embedding alignment using constraint optimization.

For completeness, we also report the accuracies achieved by state-of-the-art {\em supervised} approaches: 
{\em $~~$ (1) RCSLS}~\cite{rcsls}: State-of-the-art supervised method for training a learning architecture based on optimizing the CSLS criteria~\cite{muse}; \\
{\em $~~$ (2) GeoMM}~\cite{geomm}: Language specific geometric rotations are learnt, and subsequently a network architecture is trained to align the languages; and \\
{\em $~~$ (3) DeMa-BME}~\cite{dens}: Provides a weakly-supervised approach for learning a Gaussian Mixture Model by characterizing the probability density between embeddings spaces.

Despite obtaining state of the art results, we emphasize that achieving the best possible accuracy was not our focus. Rather, we aimed to build a framework robust to adversarial instability and data noise. Most parameters were set to fixed values. As such, following~\citet{muse}, we only fed the adversarial discriminator with the 50K most frequent words; the discriminator had an input dropout layer with rate 0.1. Production deployments may consider further parameter tuning. In our experiments, we only tuned the weight assigned to the cyclic loss between 5 and 10, and ran the framework under different random seeds, always picking the best model using the unsupervised criterion.

\subsection{Results and Discussion}

{\bf Word Translation.} Similar to machine translation, this task involves the retrieval of the translation of a given source word for a target language (from the target vocabulary). Observe, {\em polysemy} of words 
and {\em hubness} in embedding space provide a significant challenge in this setting. We evaluate the approaches using a similar setting and the ground-truth dictionaries from~\citet{muse}. From Table~\ref{tab:wt_res}, 
we observe that our $\actg$ framework provides state-of-the-art translation results in nearly all of the language pairs. In fact, for certain language pairs like {\em fr$\rightarrow$en}, the performance of $\actg$ is 
almost at par with existing supervised methods ($83.8$ compared to $84.1$ by RCSLS).

However, the challenges in word translation are compounded for {\em morphologically rich} and {\em low-resources languages} due to high vocabulary variation and limited accuracy of word embeddings respectively. To this end, 
we explore the performance of the competing algorithms on Finnish, Hebrew and Romanian --  generally identified as ``difficult'' languages in the literature~\cite{limi}. From Table~\ref{tab:wt_low} it can be seen that 
$\actg$ significantly outperforms the existing approaches with an accuracy improvement of $1.5\%$ on average across the languages.

\begin{table}[t]
\vspace*{-5mm}
\centering
\scriptsize
\caption{CSLS@1 results on well-resourced languages for the dataset of~\citet{muse}.}
\vspace*{-3mm}
\label{tab:wt_res}
	\begin{tabular}{l|cccccccc}
		\toprule
		\multirow{2}{*}{\bf Algorithm} & \multicolumn{2}{c}{\bf en-es} & \multicolumn{2}{c}{\bf en-de} & \multicolumn{2}{c}{\bf en-fr} & \multicolumn{2}{c}{\bf en-ru} \\
		\cmidrule(lr){2-3}
		\cmidrule(lr){4-5}
		\cmidrule(lr){6-7}
		\cmidrule(lr){8-9}
		& {$\rightarrow$} & {$\leftarrow$} & {$\rightarrow$} & {$\leftarrow$} & {$\rightarrow$} & {$\leftarrow$} & {$\rightarrow$} & {$\leftarrow$} \\
		\hline
		\hline
		\multicolumn{9}{c}{\it Supervised Approaches} \\
		\hline
		{\bf Non-Adv}~\cite{nonadv} & 81.4 & 82.9 & 73.5 & 72.4 & 81.1 & 82.4 & 51.7 & 63.7 \\
		{\bf DeMa-BME}~\cite{dens} & 82.8 & 85.4 & 77.2 & 75.1 & 83.2 & 83.5 & 49.2 & 63.6 \\
		{\bf GeoMM}~\cite{geomm} & 81.4 & 85.5 & 74.7 & 76.7 & 82.1 & 84.1 & 51.3 & {\em 67.6} \\
		{\bf RCSLS}~\cite{rcsls} & {\em 84.1} & {\em 86.3} & {\em 79.1} & {\em 76.3} & {\em 83.3} & {\em 84.1} & {\em 57.9} & 67.2 \\
		\hline
		\multicolumn{9}{c}{\it Unsupervised Approaches} \\
		\hline
		{\bf SinkHorn}~\cite{sinkhorn}$^{**}$ & 79.5 & 77.8 & 69.3 & 67.0 & 77.9 & 75.5 & - & - \\
		{\bf Non-Adv}~\cite{nonadv} & 82.1 & 84.1 & 74.7 &73.0 & 82.3 & 82.9 & 47.5 & 61.8 \\
		{\bf Was-Proc}~\cite{wsproc} & 82.8 & 84.1 & 75.4 & 73.3 & 82.6 & 82.9 & 43.7 & 59.1 \\
		{\bf GW-Proc}~\cite{gwproc} & 81.7 & 80.4 & 71.9 & 72.8 & 81.3 & 78.9 & 45.1 & 43.7 \\
		{\bf MUSE}~\cite{muse} & 81.7 & 83.3 & 74.0 & 72.2 & 82.3 & 82.1 & 44.0 & 59.1 \\
		{\bf VecMap}~\cite{vecmap}$^{\dag \dag}$ & 82.3 & 84.7 & 75.1 & 74.3 & 82.3 & {\em 83.6} & {\em 49.2} & {\em 65.6} \\
		{\bf UMH}~\cite{umh} & 82.5 & 84.9 & 74.8 & 73.7 & {\bf 82.9} & 83.3 & 45.3 & 62.8 \\
		{\bf Adv-Auto}~\cite{coli2020} & {\em 83.0} & {\bf 85.2} & {\bf 76.2} & {\em 74.7} & 82.3 & 83.5 & 47.6 & - \\
		\hline
		{\bf \actg} & {\bf 83.1} & {\em 85.0} & {\em 75.7} & {\bf 75.2} & {\em 82.4} & {\bf 83.8} & {\bf 49.5} & {\bf 66.1} \\
		\bottomrule
		\multicolumn{9}{l}{\scriptsize $^{**}$ Uses cosine similarity instead of CSLS and failed to reasonably converge for {\em en-ru} as reported in~\citet{dens}} \\
		\multicolumn{9}{l}{\scriptsize $^{\dag \dag}$ Results taken from~\citet{dens}} \\
	\end{tabular}
\end{table}

\begin{table}[t]
\vspace*{-2mm}
\begin{minipage}{.5\linewidth}
\scriptsize
	\caption{\small CSLS@1 results on morphologically rich and low-resource languages for~\citet{muse} data.}
	\label{tab:wt_low}
	\vspace*{-3mm}
	\begin{tabular}{l|cccccc}
		\toprule
		\multirow{2}{*}{\bf Algorithm} & \multicolumn{2}{c}{\bf en-fi} & \multicolumn{2}{c}{\bf en-he} & \multicolumn{2}{c}{\bf en-ro} \\
		\cmidrule(lr){2-3}
		\cmidrule(lr){4-5}
		\cmidrule(lr){6-7}
		& {$\rightarrow$} & {$\leftarrow$} & {$\rightarrow$} & {$\leftarrow$} & {$\rightarrow$} & {$\leftarrow$} \\
		\hline
		\hline
		{\bf MUSE} & 43.7 & 53.7 & 36.9 & - & 57.8 & 66.0 \\
		{\bf VecMap} & {\bf 49.9} & 63.1 & 44.6 & 57.5 & 64.2 & 71.8 \\
		{\bf Adv-Auto} & 49.8 & 65.7 & 46.1 & 58.6 & 61.8 & 71.9 \\
		{\bf \actg} & 49.7 & {\bf 67.3} & {\bf 46.3} & {\bf 59.1} & {\bf 65.4} & {\bf 74.3} \\
		\bottomrule
	\end{tabular}
\end{minipage}
\hspace*{3mm}
\begin{minipage}{.5\linewidth}
\scriptsize
	\caption{\small CSLS@1 results for {\em limited vocabulary} word translation on~\citet{muse} data.}
	\label{tab:wt_vocab}
	\vspace*{-3mm}
	\begin{tabular}{l|cccccc}
		\toprule
		\multirow{2}{*}{\bf Algorithm} & \multicolumn{2}{c}{\bf en-de} & \multicolumn{2}{c}{\bf en-fi} & \multicolumn{2}{c}{\bf en-ro} \\
		\cmidrule(lr){2-3}
		\cmidrule(lr){4-5}
		\cmidrule(lr){6-7}
		& {$\rightarrow$} & {$\leftarrow$} & {$\rightarrow$} & {$\leftarrow$} & {$\rightarrow$} & {$\leftarrow$} \\
		\hline
		\hline
		{\bf MUSE} & 71.0 & 77.5 & - & 71.7 & 72.7 & 75.5 \\
		{\bf VecMap} & 72.5 & 78.4 & 62.4 & 76.7 & 77.2 & 78.9 \\
		{\bf Adv-Auto} & - & - & - & - & - & - \\
		{\bf \actg} & {\bf 72.8} & {\bf 79.4} & {\bf 60.0} & {73.3} & {76.3} & {\bf 80.7} \\
		\bottomrule
	\end{tabular}
\end{minipage}
\end{table}

\comment{
{\bf Sentence Translation.} We now explore a higher level abstraction of multi-lingual word embedding space alignment, and study sentence translation retrieval on the Europarl corpus. Similar to the 
setup in~\citet{muse}, a sentence is represented as a bag-of-words, and the idf-weighted average of word embeddings are considered as the sentence embeddings. For each source sentence, the closest sentence 
(in terms of embedding distance) from the target language space is considered as its translation. We compare the precision-at-1 (P@1) results of $\actg$, on three language pairs, with the top-3 current 
state-of-the-art methods. Table~\ref{tab:sent_tran} XXXXX
}

{\bf Limited Vocabulary.} An interesting application for cross-lingual word embedding alignment is translation tasks in {\em domain-specific context}. For example, an organization expanding its scale of 
operations to geographically distributed markets and consumers. This would necessitate the efficient expansion of supporting languages for existing documents like manuals, FAQs, etc. as well as for customer 
services like Chatbots~\cite{chatbot1, chatbot2}. Observe, that in such cases, the domain-specific vocabulary is relatively small, depending on the organization's range of business range and limited training 
resources. We simulate such application scenario in this setting, and observe the performances of the algorithms in face of with limited vocabulary.

The input mono-lingual word embeddings are limited to the 10K most frequent words (instead of 200K most frequent words) in each of the languages, which can potentially severely impact the training stages of 
existing techniques. However, we initially study the {\em word coverage} of Wiki articles with varying vocabulary sizes. Figure~\ref{fig:cover} depicts the percentage of word coverage with varying frequent 
word vocabulary sizes using the plain texts of Wikipedia articles from 2018~\cite{wiki}. It can be observed, that all the languages depict similar characteristics, with a plateau around 50-75K vocabulary size. 
Recall, that our empirical setting is based on training the architectures on the $50K$ most frequent word embeddings. However, with around $10K$ vocabulary size, the coverage is in general not overtly bad 
(around 5-10\% lower), but provides valuable insights as to the robustness to domain-specific or niche applications, and is hence used in this {\em low vocabulary} setting. From Table~\ref{tab:wt_vocab}, we see that $\actg$ performs better than existing methods on most of the language pairs.

\begin{table}[t]
\vspace*{-2mm}
\begin{minipage}{0.5\linewidth}
\scriptsize
		\caption{Ablation and Robustness Study: Effect of the different modules on the overall word translation performance of \actg.}
		\label{tab:abla}
		\vspace*{-3mm}
		\begin{tabular}{l|cccccc}
		\toprule
		\multirow{2}{*}{\bf Algorithm} & \multicolumn{2}{c}{\bf en-de} & \multicolumn{2}{c}{\bf en-fi} & \multicolumn{2}{c}{\bf en-ro} \\
		\cmidrule(lr){2-3}
		\cmidrule(lr){4-5}
		\cmidrule(lr){6-7}
		& {$\rightarrow$} & {$\leftarrow$} & {$\rightarrow$} & {$\leftarrow$} & {$\rightarrow$} & {$\leftarrow$} \\
		\hline
		\hline
		{\bf MUSE GAN} & 70.1 & 66.4 & 22.3 & 24.1 & 34.5 & 49.6 \\	
		{\bf CycleGAN} & 71.2 & 70.7 & 28.7 & 48.7 & 43.5 & 48.7 \\
		{\bf CycleGAN} & 71.2 & 70.7 & 28.7 & 48.7 & 43.5 & 48.7 \\
		{\bf CycleGAN + Sym-Wh.} & 75.5 & 74.9 & 47.9 & 66.1 & 63.8 & 72.5 \\
		{\bf \actg} & {\bf 75.7} & {\bf 75.2} & {\bf 49.7} & {\bf 67.3} & {\bf 65.4} & {\bf 74.3} \\
		\hline
		{\bf Bad-GAN} & 57.7 & 66.7 & 27.0 & 31.1 & 37.9 & 46.8 \\
		{\bf $\actg$ with Bad-GAN} & 75.1 & 75.3 & 50.9 & 66.3 & 64.3 & 73.7 \\
		\bottomrule
	\end{tabular}
	\end{minipage}
	\hspace*{15mm}
	\begin{minipage}{0.5\linewidth}
		\includegraphics[width=55mm]{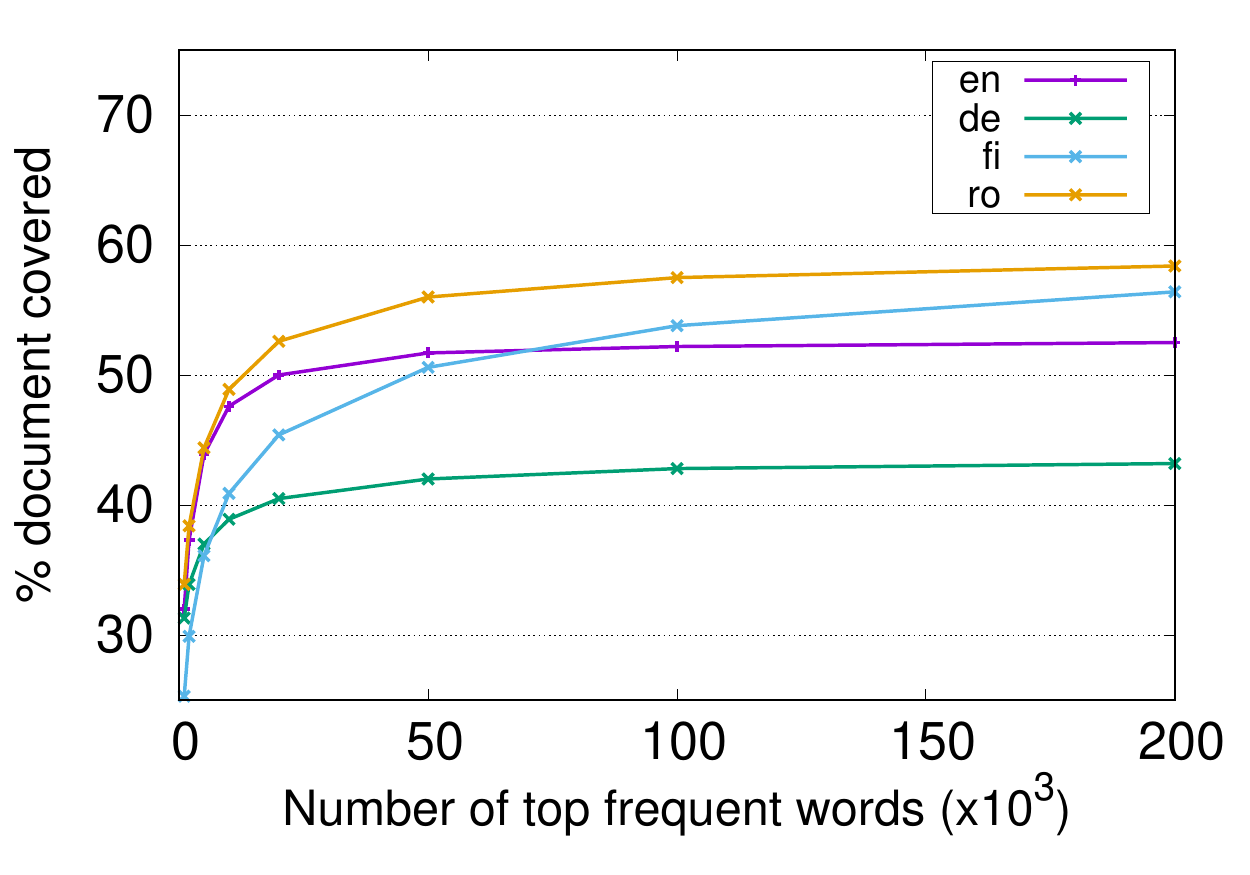}
		\vspace*{-3mm}
		\captionof{figure}{\small Distribution of document \\ coverage with truncated vocabulary set.}
		\label{fig:cover}
	\end{minipage}
\end{table}

{\bf Ablation Study.} Finally, to understand the effects of the different modules in $\actg$ on the overall performance, we perform ablation study by incrementally adding and removing the separate components.
Table~\ref{tab:abla} tabulates the obtained results on different language pairs (including morphologically rich and low-resourced). We observe, that the adversarial network, CycleGAN, using the cycle-loss 
consistency optimization criteria, in general performs better than MUSE GAN, the traditional GAN framework of~\citet{muse}. In terms of refinement performed in the {\em Correspond} module of $\actg$, we compared 
the performance of symmetric re-whitening (used in this work) with the orthogonal Procrustes strategy. Both the refinement processes are seen to be comparable in performance, however since Procrustes, by definition, 
is a supervised approach, errors from the adversarial training in the {\em Align} module might be propagated, degrading the efficacy of the entire framework. Finally, addition of the Coherent Point Drift point-set 
registration in the {\em Transform} module (i.e., the complete $\actg$ pipeline) is seen to further improve the results over the refinement strategy.

One important criticism for the performance adversarial training based alignment techniques is their dependence on precise parameter settings to tackle convergence instability (reported previously in our empirical results). 
Hence, we study the {\em robustness} of $\actg$ to such issues, by intentionally selecting a sub-optimal CycleGAN model (from the training epochs) as the final output from the adversarial based {\em Align} module, denoted 
as {\em Bad-GAN} in Table~\ref{tab:abla}. We observe $\actg$ to robustly handle such situations, and provide a final accuracy score that is comparable to that achieved with a properly 
trained adversarial model selected based on our {\em cyclic unsupervised criteria}. Specifically, for {\em en $\rightarrow$ de} and {\em fi $\rightarrow$ en} languages, the performance of Bad-GAN is around $15\%$ 
worse than the properly selected CycleGAN model, however, the final accuracy of $\actg$ for word translation is seen to differ by only $1\%$ (Table~\ref{tab:abla}) -- depicting robustness to noisy training.

In summary, the above empirical evaluations showcase that our framework, $\actg$, provides better unsupervised cross-lingual alignment of embedding spaces, by not only outperforming existing techniques in terms 
of translation accuracy even on morphologically rich and low-resource languages, but also demonstrating robustness in gracefully handling potential adversarial training loss.
\section{Conclusion}
\label{sec:conc}
We introduced $\actg$, an unsupervised cross-lingual alignment framework for word embedding. We use {\em }adversarial training with a cycle-consistency loss to induce a seed bidirectional mapping, that we subsequently refine and generate word correspondences using {\em point set registration method}. Extensive experiments on multiple languages for parallel dictionary creation not only demonstrate state-of-the-art results for our framework, but also depict robustness to variable adversarial performance, a considerable limitation of past work.

\end{document}